\documentclass[runningheads]{llncs}
\usepackage[T1]{fontenc}
\usepackage{amsmath,amssymb, amsfonts, bbm}
\usepackage{graphicx,verbatim}

\usepackage{ algorithmic, algorithm}
\usepackage{multirow}
\usepackage{booktabs, graphicx}
\usepackage{tabularray}
\UseTblrLibrary{booktabs}
\usepackage{hyperref}
\usepackage{color}

\begin{document}
\title{Deep EM with Hierarchical Latent Label Modelling for Multi-Site Prostate Lesion Segmentation}
\titlerunning{HierEM for Multi-Site Prostate Lesion Segmentation}
%


\author{Wen Yan\inst{1}, Yipei Wang\inst{1}, Shiqi Huang\inst{1}, Natasha Thorley\inst{2}, Mark Emberton\inst{3}, Vasilis Stavrinides\inst{4,5}, Shonit Punwani\inst{2}, Yipeng Hu\inst{1}, Dean Barratt\inst{1}}  
%

\authorrunning{Wen Yan et al.}
\institute{$^1$ UCL Hawkes Institute; Department of Medical Physics and Biomedical Engineering, University College London, Gower St, London WC1E 6BT, U.K. \\
    \email{wen-yan@ucl.ac.uk}\\                                  $^2$ Centre for Medical Imaging, Division of Medicine, University College London,  Gower St, London WC1E 6BT, U.K.\\
    $^3$ Division of Surgery and Interventional Science, University College London, 10 Pond St, London NW3 2PS, U.K.\\
    $^4$ Cancer Institute, Urology Department, UCLH, University College London, 16-18 Westmoreland St, London W1G 8PH, U.K.\\
$^5$ Radiology Department, Imperial College Healthcare,  The Bays, S Wharf Rd, London W2 1NY, U.K.}
  
\maketitle              
\begin{abstract}
Label variability is one of the major challenges in prostate lesion segmentation. In multi-site datasets, annotations often reflect site-specific contouring protocols. This can bias segmentation networks towards local annotation styles and reduce generalisation to unseen sites during inference. In this study, we treat each observed annotation as a noisy observation of an underlying latent ``clean'' lesion mask, and propose a hierarchical expectation–maximisation (HierEM) framework that alternates between: (i) inferring a voxel-wise posterior distribution over the latent mask, and (ii) training a CNN using this posterior as a soft target and estimating site-specific sensitivity and specificity using a logistic-normal hierarchical prior. The hierarchical prior decomposes label quality into global, site-, and case-level components, which regularises site-level deviations, encouraging the model to avoid overfitting to centre-specific annotation styles, thereby promoting site-invariant segmentation. 
Experiments on three-site datasets demonstrate that the proposed hierarchical EM framework enhances cross-site generalisation compared to state-of-the-art methods.
In the pooled datasets evaluation, per-site mean DSC ranges from 29.50\% to 39.69\% across datasets; for leave-one-site-out generalisation, it ranges from 27.91\% to 32.67\%, yielding statistically significant improvements over comparison methods ($p<0.039$). The method also produces interpretable per-site latent label-quality estimates (e.g., sensitivity $\alpha$ ranging from 31.5\% to 47.3\% at specificity $\beta\approx0.99$). These results indicate that explicitly modelling site-dependent annotation can improve cross-site generalisation. Code is available via \url{https://github.com/yanwenCi/HierEM}.

\end{abstract}

\section{Introduction}
Multiparametric MRI (mpMRI), including T2-weighted (T2W), diffusion-weighted imaging (DWI), apparent diffusion coefficient (ADC) maps and contrast-enhanced (DCE) modalities, is widely used to assess suspected prostate cancer (PCa)~\cite{haider2007combined}. Clinicians often contour lesions to guide diagnosis, treatment planning and follow-up.
Automated deep-learning methods aim to reduce this workload, yet they face a significantly high-variance ground truth challenge. 
This label variability is rooted in site-specific annotation protocols, institutional contouring styles shaped by local expert training, and imaging protocols that influence how lesion boundaries are perceived. Together, these factors create site-specific label bias.
Inter-reader agreement for prostate lesion contours on mpMRI is only moderate, with reported Dice often around $\sim0.4$~\cite{liechti2020manual,piert2018accuracy}. 
The current paradigm creates a critical bottleneck: segmentation networks easily overfit to the local contouring style of the training sites. When deployed to a new institution, the model generalises poorly, as single-site training achieves only 4\%–28\% Dice on a held-out site~\cite{rodrigues2024prostatnet}. By contrast, testing on within-site achieves around $20-60\%$ Dice~\cite{yan2024combiner,hambarde2020prostate,wang2024paracm,fouladi2026exploring} with different datasets.
This gap can be alleviated via test-site finetuning or calibration. However, such methods often lead to test-site-biased accuracy assessments. They force the model to match the local ``observed'' labels, which are themselves imperfect and biased. A truly multi-site training set (more than 10 sites) is expensive to curate for individual tasks that require coordination in labelling. PI-CAI is an example study with such a diverse collection of reference standards, but restricted to a primarily patient-level classification~\cite{saha2023artificial}. In many real-world applications, further finetuning or calibration at deployment may not be feasible or practical.
While cross-site performance gaps may also stem from both imaging differences and label variation, pre-processing can partially mitigate the former. In this study, we focus on the impact of label-level variation across sites. 

We propose a latent-label modelling approach to address cross-site annotation variability using site-level sensitivity and specificity parameters. Our use of sensitivity and specificity is related to the classical formulation used in STAPLE~\cite{warfield2004simultaneous}, but the problem setting is fundamentally different. STAPLE addresses multi-annotator label fusion for a fixed image, where multiple annotations of the same case are available and the goal is to infer a consensus label while estimating annotator reliability. In contrast, our setting typically contains only one annotation per case, and the annotation source is determined by the acquisition site rather than by multiple independent raters. We therefore do not perform label fusion. Instead, we model each observed site-specific annotation as a noisy label of an unobserved latent lesion mask during training. The segmentation network estimates this latent mask, while site-dependent sensitivity and specificity parameters account for systematic differences in annotation behaviour across sites. Thus, sensitivity and specificity are used not for multi-rater fusion, but to characterise site-dependent label noise and support learning of a site-agnostic lesion representation from single, site-dependent noisy labels.
We impose logistic-normal hierarchical priors on the sensitivity and specificity parameters, which are decomposed into three components: (1) global population factors capturing lesion characteristics shared across sites; (2) site-specific effects modelling systematic offsets due to local contouring and imaging protocols; and (3) case-level variability capturing intrinsic ambiguity, such as small or low-contrast lesions, that may affect annotation independently of site protocol. This decomposition enables partial pooling across multi-site data while accounting for both systematic site-level variation and individual-case ambiguity.

Learning is performed with an EM procedure~\cite{moon1996expectation} that alternates between inferring the latent consensus mask and updating model parameters. In the E-step, we compute a voxel-wise posterior over the latent consensus mask by combining the network’s image-based prior with the annotation likelihood under the current model. In the M-step, we update the UNet using this posterior as soft targets, and re-estimate the site-/case-level sensitivity and specificity by maximising their marginal likelihood under the hierarchical prior. Iterating these steps reduces dependence on any single site’s contouring style, yields calibrated probabilistic predictions, and provides interpretable sensitivity and specificity estimates that summarise site tendencies and case-level ambiguity—improving robustness when deploying to unseen sites without requiring additional site-specific fine-tuning.

\section{Method}
\label{sec:method}

Let $\Omega$ denote the voxel space of an mpMRI volume. For each case $k\in\{1,\dots,K\}$ we observe an image $X_k$ and a single binary lesion annotation $Y_k\in\{0,1\}^\Omega$ provided by one site $s_k\in\{1,\dots,S\}$. The (unobserved) latent ``clean'' lesion mask is $G_k\in \{0,1\}^\Omega$.
Our goal is to learn a segmentation model that is robust to site-specific annotation protocol, while also estimating site-/case-level sensitivity and specificity for observed labels relative to the latent true labels.

\begin{figure}
    \centering
    \includegraphics[width=\linewidth]{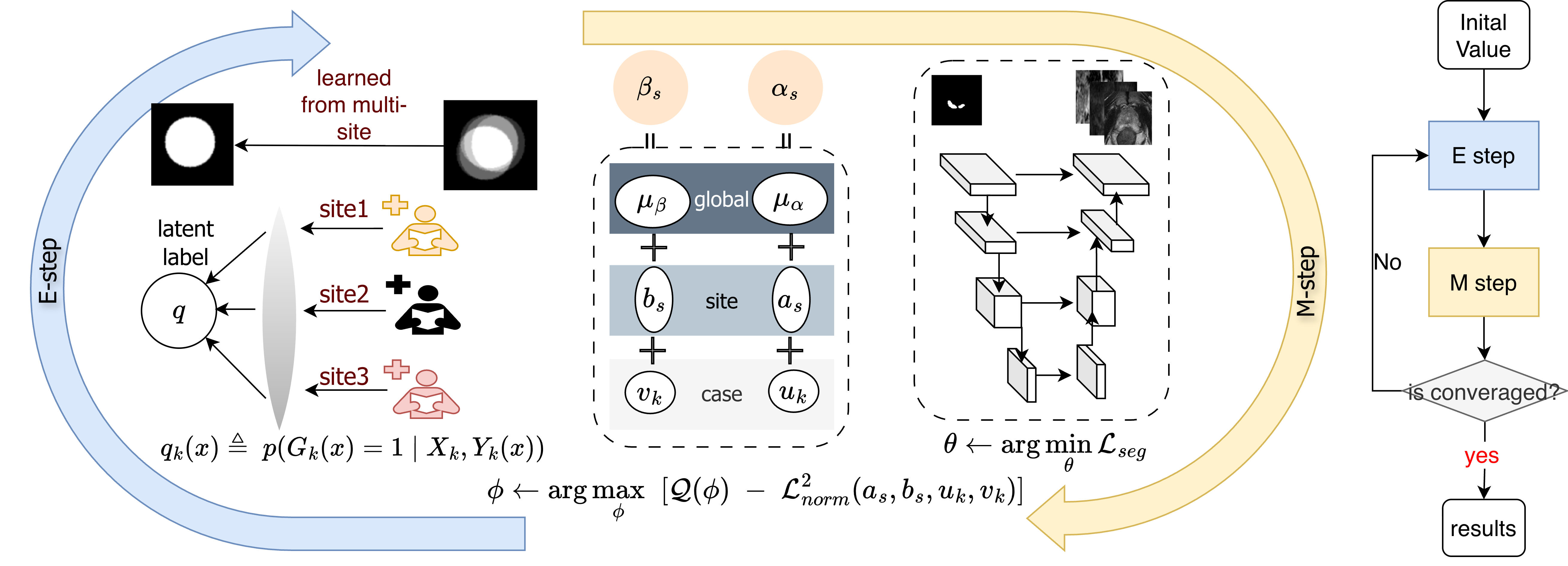}
    \caption{Overview of proposed deep mixed EM framework with hierarchical latent label-quality parameters. The E-step infers a latent clean mask, and the M-step updates the segmentation network and site-specific sensitivity and specificity.}
    \label{fig:main}
\end{figure}

\paragraph{Segmentation network}
We model the latent consensus mask with an image-conditioned voxel-wise probability map predicted by the network:
\(
\pi_k(x) \;\triangleq\; p_\theta\!\left(G_k(x)=1 \mid X_k\right)
\;=\; \sigma\!\left(f_\theta(X_k)_x\right),
\)
where $x \in \Omega$ is a voxel in mpMRI, $f_\theta(\cdot)$ is a neural network producing logits per voxel and $\sigma(\cdot)$ is the sigmoid.

\subsection{Hierarchical site-case label variability model}
We treat each site as having a distinct annotation protocol and model the observed label $Y_k(x)$ as a noisy observation of $G_k(x)$.
Conditioned on $(G_k, s_k)$, voxels are assumed independent with site- and case-level sensitivity and specificity:
\begin{align}
p(Y_k(x)=1 \mid G_k(x)=1) = \alpha_{s_k,k},\quad
p(Y_k(x)=0 \mid G_k(x)=0) = \beta_{s_k,k},\ \forall x\in\Omega_k.
\end{align}
To stabilise site-/case-level label-quality estimates, we use a logistic-normal hierarchical prior that partially pools sensitivity and specificity toward a global mean $\mu_\alpha, \mu_\beta$:
\begin{equation}
\text{logit}(\alpha_{s,k}) = \mu_\alpha + a_s + u_k, \qquad
\text{logit}(\beta_{s,k})  = \mu_\beta + b_s + v_k, 
\label{eq:beta_hier}
\end{equation}
where $(a_s,b_s)$ capture persistent site-specific contouring behaviour and $(u_k,v_k)$ capture case difficulty.
We place zero-mean Gaussian priors (equivalently, $\ell_2$ penalties under MAP): \(a_s\sim\mathcal N(0,\sigma_a^2),\;\; b_s\sim\mathcal N(0,\sigma_b^2),\;\;
 u_k\sim\mathcal N(0,\sigma_u^2),\;\; v_k\sim\mathcal N(0,\sigma_v^2)\). 
For identifiability we constrain $\sum_s a_s = 0, \quad \sum_s b_s = 0$. We further calculate site-level sen/spec as ($\sigma(\cdot)$ is sigmoid function):
\begin{equation}
    \alpha_s = \sigma(\mu_\alpha+a_s), \qquad \beta_s = \sigma(\mu_\beta+b_s)
\end{equation}

\textbf{EM learning with hierarchical sensitivity/specificity modelling}
We estimate segmentation network parameters $\theta$ and the latent label-quality parameters
$\phi=\{\mu_\alpha,\mu_\beta,a_{1:S},b_{1:S},u_{1:K},v_{1:K}\}$
by maximizing the (regularized) marginal likelihood of $\{Y_k\}$ with latent $\{G_k\}$.
We optimise a MAP objective via an EM procedure.

\paragraph{E-step: posterior of the latent mask.}
Given current $(\theta,\phi)$, the posterior factorizes over voxels:
\begin{align}
q_k(x) &\;\triangleq\; p(G_k(x)=1 \mid X_k, Y_k(x)) \\
&\;=\;
\frac{\pi_k(x)\,p(Y_k(x)\mid G_k(x)=1)}{\pi_k(x)\,p(Y_k(x)\mid G_k(x)=1) + (1-\pi_k(x))\,p(Y_k(x)\mid G_k(x)=0)}.
\label{eq:posterior_q}
\end{align}
With the sensitivity and specificity model, the likelihood terms are
\begin{align}
&p(Y\!=\!1 \mid G\!=\!1)=\alpha_{s,k},\quad p(Y\!=\!1 \mid G\!=\!0)=1-\beta_{s,k},\quad
\\
& p(Y\!=\!0 \mid G\!=\!1)=1-\alpha_{s,k},\quad p(Y\!=\!0 \mid G\!=\!0)=\beta_{s,k}.
\end{align}
\begin{equation}
    p(Y|G;\alpha, \beta)=
    \begin{cases}
        \alpha^Y(1-\alpha)^{1-Y}, \qquad G=1, \\
        (1-\beta)^Y\beta^{1-Y},\qquad G=0. \\ 
    \end{cases}
    \label{eq:log-like}
\end{equation}
Thus $q_k(x)$ provides a \emph{soft consensus} mask that fuses the image prior and the site- and case-level sensitivity and specificity.

\paragraph{M-step (A): update segmentation network.}
We update the segmentation network with $\theta$ by minimising a soft-label cross-entropy plus Dice loss:
\begin{equation}
\mathcal L_{\text{seg}}(\theta) \;=\; \sum_{k=1}^K \sum_{x\in\Omega}
\mathrm{CE}\big(q_k(x), \pi_k(x)\big) 
- \sum_{k=1}^{K}\mathrm{Dice}(q_k(x),\pi_k(x))
,
\label{eq:soft_ce}
\end{equation}
where $\mathrm{CE}(q,\pi) = -q\log \pi -(1-q)\log(1-\pi)$ and  \(\mathrm{Dice}(q,\pi)=\frac{ \sum_{x\in \Omega}2 q_k(x)\cdot\pi_k(x)}{\sum_{x\in \Omega} q_k(x)+\sum_{x\in \Omega} \pi_k(x)}\)

\paragraph{M-step (B): update hierarchical latent label-quality by aggregated expected counts.}
Define expected per-case sufficient statistics:
\begin{align}
TP_k &= \sum_{x\in\Omega} q_k(x)\,Y_k(x), \qquad &P_k &= \sum_{x\in\Omega} q_k(x), \label{eq:suff_tp}\\
TN_k &= \sum_{x\in\Omega} (1-q_k(x))\,(1-Y_k(x)), \qquad &N_k &= \sum_{x\in\Omega} (1-q_k(x)). \label{eq:suff_tn}
\end{align}
Using posterior $q$ to approximate latent Ground Truth $G$, the expected complete-data log-likelihood for latent label-quality parameters is:
\begin{align}
\mathcal Q(\phi)
& = \mathbb{E}_{G\sim q}\big[\log p(Y|G;\phi)\big]\\
& =\sum_{x\in\Omega}\Big[\mathbbm{1}\!\left\{G_k(x)=1\right\}\Big(Y_k(x)\log \alpha_{s_k,k} + \big(1-Y_k(x)\big)\log\big(1-\alpha_{s_k,k}\big)\Big)
\nonumber\\
&+\mathbbm{1}\!\left\{G_k=0\right\}
\Big(\big(1-Y_k(x)\big)\log \beta_{s_k,k} + Y_k(x)\log\big(1-\beta_{s_k,k}\big)\Big)
\Big]\\
&=\sum_{k=1}^K \Big[
TP_k \log \alpha_{s_k,k} + (P_k-TP_k)\log(1-\alpha_{s_k,k}) \nonumber\\
&+ TN_k \log \beta_{s_k,k} + (N_k-TN_k)\log(1-\beta_{s_k,k})
\Big],
\label{eq:Q_phi}
\end{align}

\noindent with $\alpha_{s,k},\beta_{s,k}$ defined by \eqref{eq:beta_hier}.
We obtain a MAP estimate by maximising MLE and L2 penalisation on prior variables in $\phi$:
\begin{equation}
\phi \leftarrow \arg\max_\phi \;\; [\mathcal Q(\phi)\;-\; \mathcal{L}^2_{norm}], 
\label{eq:map_phi}
\end{equation}
where $\mathcal{L}^2_{norm}=\frac{1}{2\sigma_a^2}\sum_s a_s^2 + \frac{1}{2\sigma_b^2}\sum_s b_s^2
+\frac{1}{2\sigma_u^2}\sum_k u_k^2 + \frac{1}{2\sigma_v^2}\sum_k v_k^2$
subject to $\sum_s a_s=0$ and $\sum_s b_s=0$.
Because \eqref{eq:map_phi} depends on the data only via $(TP_k,P_k,\\TN_k,N_k)$, it can be optimised efficiently with a few iterations of a second-order L-BFGS~\cite{nocedal1980updating} over low-dimensional parameters. $L_2$ penalty on the hierarchical latent variables in $\phi$ corresponds to a Gaussian prior and provides shrinkage that stabilises site/case quality estimates, prevents degenerate sensitivity and specificity values, and mitigates overfitting when annotations are sparse or noisy.

\paragraph{Training procedure}\label{sec:train}
We alternate the E-step \eqref{eq:posterior_q} with M-steps \eqref{eq:soft_ce} and \eqref{eq:map_phi} until convergence. We used the Adam optimiser with a learning rate of $10^{-4}$. 
In practice, we perform 5-epoch gradient steps for $\theta$ per EM iteration, and 5 optimiser steps for $\phi$ using the aggregated sufficient statistics in~\eqref{eq:Q_phi}.
Algorithm~\ref{alg:hier_em} summarises the procedure. To stabilise early optimisation, we add an auxiliary supervision in ~\eqref{eq:soft_ce} using the observed label. The corresponding loss weight was gradually decayed to zero over the first five epochs using a sigmoid schedule.

\begin{algorithm}[t]
\caption{Hierarchical EM for multi-site lesion segmentation}
\label{alg:hier_em}
\begin{algorithmic}[1]
\REQUIRE Images $\{X_k\}$, masks $\{Y_k\}$, site labels $\{s_k\}$
\STATE Initialize network $\theta$; initialize $\phi$ ( $\mu_\alpha,\mu_\beta$ near $\text{logit}(0.9)$, $a_s=b_s=u_k=v_k=0$)
\FOR{EM iterations $t=1,\dots,T$}
    \STATE \textbf{E-step:} compute $\pi_k(x)=\sigma(f_\theta(X_k)_x)$ and $q_k(x)$ via \eqref{eq:posterior_q}
    \STATE Compute $(TP_k,P_k,TN_k,N_k)$ via \eqref{eq:suff_tp}--\eqref{eq:suff_tn}
    \STATE \textbf{M-step (A):} update $\theta$ by minimizing $\mathcal L_{\text{seg}}(\theta)$ in \eqref{eq:soft_ce}
    \STATE \textbf{M-step (B):} update $\phi$ by MAP optimization of \eqref{eq:map_phi} with constraints $\sum_s a_s=0$, $\sum_s b_s=0$
\ENDFOR
\STATE \textbf{Output:} network $\theta$ and learned site-/case-level parameters $\phi$
\end{algorithmic}
\end{algorithm}

\subsection{Uncertainty and interpretability}
We quantify voxel-wise uncertainty using the predictive entropy of the segmentation probability map.
The entropy at voxel $i$ is
\(
H_i \;=\; -p_i\log(p_i) \;-\; (1-p_i)\log(1-p_i).
\)
Entropy is small when $p_i$ is close to $0$ or $1$ (confident predictions) and maximised at $p_i=0.5$.
To evaluate whether uncertainty aligns with segmentation errors, we construct a risk-coverage curve by ranking voxels by entropy and retaining the lowest-entropy fraction $c\in[0.5,0.9]$.
For each $c$, we choose a threshold $t_c$ such that
$
S_c \;=\; \{\, i \;:\; H_i \le t_c \,\},  \frac{|S_c|}{|\Omega|}=c,
$
where $\Omega$ denotes the set of all voxels.
We then compute the Dice score restricted to the retained set $S_c$, denoted $\mathrm{Dice}(S_c)$, and define risk as
\(
\mathrm{Risk}(c) \;=\; 1 - \mathrm{Dice}(S_c).
\)

\section{Experiment}
\label{sec:results_plan}

\subsection{Datasets}
We utilised datasets from $S=3$ sites. Each site provides T2W, DWI$_{\mathrm{high}\,b}$, and ADC images, together with a single expert lesion contour annotated on T2W. Sites~1 was obtained from UCLH with ethical approval, described in~\cite{yan2024combiner}. The dataset comprises 850 patients of 1201 mpMR combinations with multiple high b-values, each with PI-RADS $\geq 3$ lesions. Site~2 is a public cohort~\cite{wang_2025_15683922}, which includes 391 patients who have Likert $\geq$3 lesions. 
Site~3 was collected under ethically approved clinical protocols across multiple institutions in Miami, FL, containing 1265 mpMRI scans with PIRADS $\geq 3$ lesion contours.

\textbf{Split A (Pooled patient-level held-out evaluation).}
We split each dataset by 4:1:1, resulting in pooled mixed train, validation and per-patient level held-out sets. The held-out sets were only used for model evaluation.

\textbf{Split B (Leave-one-site-out, LOSO, cross-site generalisation).}
We perform leave-one-site-out evaluation across the 3 sites. For each fold, we hold out all cases from one site as test,
and train on the remaining two sites. Within the training pool, we reserve $10\%$ of cases (stratified by site) as validation
for early stopping and hyperparameter tuning.


\subsection{Baselines and evaluation metrics}

\paragraph{Baselines} In this study, to ensure a fair and transparent comparison across methods, we use UNet~\cite{ronneberger2015u} with nnUNet-derived preprocessing and hyperparameters~\cite{isensee2021nnu} as the segmentation backbone for the proposed method and all compared methods. We compare against:
(i) Supervised UNet;
(ii) label bootstrapping (soft self-training)~\cite{reed2014training}, where the standard supervised loss is replaced by a bootstrapped self-training loss that combines the provided labels with the model's own predictions as soft auxiliary targets:
\(
\mathcal{L}_{bootstrap}^{t+1}
=
(1-\lambda)\mathcal{L}_{seg}(\omega^{t+1}, y)
+
\lambda\mathcal{L}_{seg}(\omega^{t+1}, \omega^{t}),
\)
where $\omega^t$ denotes the network prediction at iteration $t$, and $\lambda$ is a ramp-up hyperparameter with a maximum value of 0.5. This encourages smoother decision boundaries and reduces sensitivity to annotation noise.
(iii) Site-as-reader EM without hierarchy, where independent $\alpha_s,\beta_s$ are estimated for each site rather than using the hierarchical formulation in Eq.~\eqref{eq:beta_hier}.
This progressive comparison serves both as a baseline comparison and as an ablation study, isolating the contribution of the key components: soft-label self-training, explicit EM-based noise modelling, and hierarchical site/case reliability modelling.

\paragraph{Evaluation metrics for segmentation and uncertainty}
On each test fold and site, we report
Dice and HD95 for segmentation results. 
For calibration and uncertainty metrics, we report selective segmentation (risk-coverage) curves based on predicted uncertainty (entropy of $\pi_k(x)$). We compare methods using a paired $t$-test on the same test cases within each held-out site.
Multiple comparisons are controlled within each held-out site using Benjamini-Hochberg FDR at 5\%.

\begin{table*}[tb]
\centering
\caption{Sites 1–3 contained 252, 87, and 265 cases in the pooled patient-level held-out set, and 1201, 391, and 1265 cases in the LOSO experiments, respectively. $*$ denote HierEM is significantly better than other baselines based on statistical tests with a significance level of $\alpha=0.05$.}
\label{tab:loso_main}

{\fontsize{8pt}{9.2pt}\selectfont
\begin{tblr}{
  width=\dimexpr\textwidth+0pt\relax,
  colspec={
    Q[l,wd=0.12\textwidth]
    *{3}{ Q[c,wd=0.15\textwidth] Q[c,wd=0.14\textwidth] }
  },
  colsep=1pt,
  rowsep=1.2pt,
  row{1-2}={font=\bfseries},
  cell{3-Z}{2-7}={mode=math},
}
\toprule
Methods
& \SetCell[c=2]{c} Site 1
& & \SetCell[c=2]{c} Site 2
& & \SetCell[c=2]{c} Site 3
& \\
\cline{2-3,4-5,6-7}
& Dice(\%) & HD95(mm) & Dice(\%) & HD95(mm) & Dice(\%) & HD95(mm) \\
\cline{2-3,4-5,6-7}
&\SetCell[c=6]{c}\textit{Split A: Pooled patient-level held-out split} \\
\midrule
UNet
& 38.63\!\pm\!10.41 & 17.04\!\pm\!2.96
& 27.47\!\pm\!9.62^*  & 21.25\!\pm\!4.50
& 33.49\!\pm\!6.98^*  & 22.06\!\pm\!3.92 \\
Bootstrap
& 38.49\!\pm\!9.94 & 16.97\!\pm\!3.14 & 26.69\!\pm\!10.59^* & 22.29\!\pm\!5.12^* & 33.14\!\pm\!7.05^* & \mathbf{21.41\!\pm\! 5.84}\\
Site-EM
& 36.46\!\pm\!11.72^* & 18.41\!\pm\!7.07^* & 23.74\!\pm\!9.73^* & 22.99\!\pm\!5.52^* & 34.62\!\pm\!6.15 & 21.86\!\pm\! 4.91\\
\textbf{HierEM}
& \mathbf{39.69}\!\pm\!\mathbf{11.88} & \mathbf{16.92}\!\pm\!\mathbf{3.29}
& \mathbf{29.50}\!\pm\!\mathbf{9.13}  & \mathbf{20.46}\!\pm\!\mathbf{5.58}
& \mathbf{35.60}\!\pm\!\mathbf{7.85} & {21.66}\!\pm\!{4.11} \\
\midrule

&\SetCell[c=6]{c}\textit{Split B: Leave-one-site-out (LOSO)} \\
\midrule
UNet
& 25.50\!\pm\!7.33^* & 22.10\!\pm\!4.65^*
& 24.66\!\pm\!5.29^* & 22.46\!\pm\!3.15^*
& 31.20\!\pm\!9.80^* &\mathbf{21.05\!\pm\!5.64} \\
Bootstrap
& 26.34\!\pm\!12.92^* & 21.02\!\pm\!7.02^* & 21.66\!\pm\!5.20^* & 22.54\!\pm\!3.56^* & 31.64\!\pm\!8.92^*  & 22.85\!\pm\!5.17 \\
Site-EM
& 24.00\!\pm\!18.09^* & 23.74\!\pm\!10.79^* & 24.92\!\pm\!13.29^* & 21.83\!\pm\!7.13 & 31.90\!\pm\!13.43^* & 22.50\!\pm\!7.28 \\
\textbf{HierEM}
& \mathbf{28.11}\!\pm\!\mathbf{10.21} & \mathbf{20.51}\!\pm\!\mathbf{4.68}
& \mathbf{27.91}\!\pm\!\mathbf{10.16}  & \mathbf{21.12}\!\pm\!\mathbf{5.83}
& \mathbf{32.67}\!\pm\!\mathbf{8.08} & {22.82}\!\pm\!{2.50} \\
\bottomrule
\end{tblr}\label{tab:compare}
}
\end{table*}

\begin{figure}[!b]
    \centering
    \includegraphics[width=\linewidth]{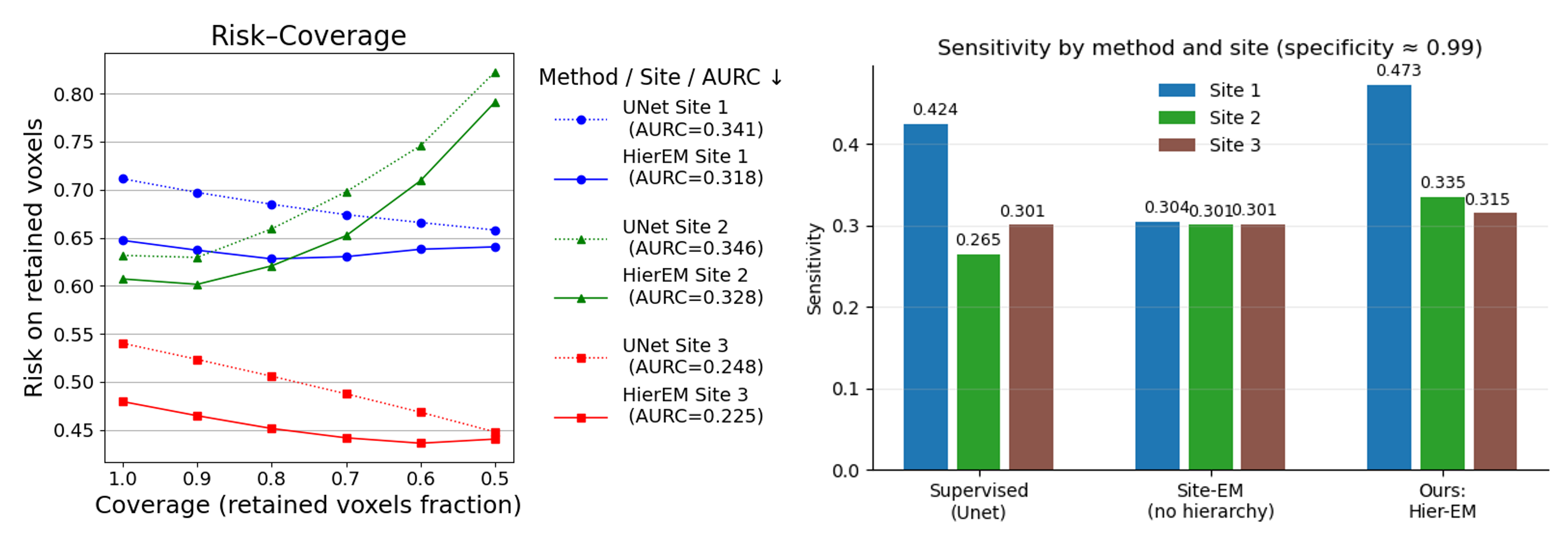}
    \caption{Left: Risk-coverage curves on three-site datasets under LOSO experiments, where smaller AURC represents lower risk; Right: Estimated sensitivity of baselines and HierEM with $spec\simeq0.99$. Such high specificity is expected because lesions occupy only a small fraction of the image.}
    \label{fig:uncertainty}
\end{figure}

\section{Results}
Table \ref{tab:loso_main} reports cross-site performance using Dice and HD95. Under the pooled patient-level held-out split, all methods achieve moderate Dice scores, approximately from 27\% to 40\%, while HierEM gives the best mean Dice on all three sites (Site 1: 39.69\%, Site 2: 29.50\%, Site 3: 35.60\%) and comparable or slightly improved HD95 (e.g., Site 2: 20.46 vs 21.25 for Supervised). Bootstrap is broadly similar to Supervised on the pooled split. In the pooled held-out split, significance is not observed for every comparison, likely due to limited test-set size. In addition, pooled multi-site training improves baseline performance, so our method shows a smaller margin in this setting. 
Under the more challenging LOSO evaluation, performance drops substantially for methods trained without explicit cross-site label variability modelling. Supervised Dice decreases to 25.50/24.66/31.20\% across Sites 1–3, with a concurrent increase in boundary error compared to pooled held-out testing. In contrast, HierEM consistently improves generalisation, yielding higher Dice on Sites 1 and 2 (28.11\% and 27.91\%) with lower HD95 (20.51 and 21.12 mm), while remaining competitive on Site 3 (Dice 32.67\%; HD95 22.82 mm). Overall, HierEM provides the most consistent cross-site performance, especially in the LOSO setting.
Fig~\ref{fig:uncertainty}-left shows risk–coverage curves that assess uncertainty for selective segmentation. HierEM has lower risks than supervised UNet at a given coverage range for all three sites LOSO experiments, which indicates that the method’s uncertainty concentrates errors in the rejected region, enabling reliable abstention. In Figure~\ref{fig:uncertainty}-right, HierEM achieved the highest sensitivity under the matched specificity ($\simeq0.99$) in all LOSO experiments. 
The gap between pooled and LOSO highlights a strong domain shift: models that perform reasonably well on pooled held-out testing degrade when the test site is unseen, suggesting that performance in pooled splits largely comes from learning site-specific contouring biases rather than a fully transferable latent ``clean'' lesion mask. HierEM narrows this gap, supporting the hypothesis that explicitly modelling site/case-dependent label noise helps separate site-specific annotation from image evidence. By inferring a soft latent consensus mask ($q$) and updating sensitivity and specificity, HierEM reduces overfitting to training-site annotation styles when applied to a new site, which is reflected in both higher Dice and lower HD95 on unseen sites.

\section{Conclusion}
We propose a Deep EM framework with a hierarchical prior that treats each site’s lesion contour as a noisy observation of a latent clean mask. The E-step infers a voxel-wise posterior over this latent mask, while the M-step updates the segmentation network and estimates site-specific sensitivity and specificity using hierarchical priors. Rather than representing individual-reader performance, the estimated sensitivity and specificity represent pooled site-level labelling behaviour, capturing the effective annotation tendency of each dataset/site. By separating latent labels from site- and case-dependent annotation variability, HierEM improves robustness and cross-site generalisation while providing interpretable diagnostics of site-level annotation behaviour. The framework is backbone-agnostic, and future work will extend it to multi-site, multi-annotator datasets and richer models of clinical annotation variability.

\begin{credits}
\subsubsection{\ackname} 
This work is supported by the International Alliance for Cancer Early Detection, an alliance between Cancer Research UK [C28070/A30912; C73666/A31378; EDDAMC-2021/100011], Canary Center at Stanford University, the University of Cambridge, OHSU Knight Cancer Institute, University College London and the University of Manchester. 

\subsubsection{\discintname}
The authors have no competing interests to declare that are relevant to the content of this article
\end{credits}

%
%
%
\bibliographystyle{splncs04}
\bibliography{mybibliography}

\end{document}